# ORDERS: An Empirical Study of Norm-Rank Aggregation for Personalized Federated Learning

Koffka Khan

Department of Computing and Information Technology, Faculty of Science and Technology, The University of the West Indies, St. Augustine, Trinidad and Tobago

koffka.khan@sta.uwi.edu

## Abstract

Personalized federated learning combines shared representations with client-specific predictors, but the contribution of a server weighting rule can be obscured by local training and evaluation choices. We study ORDERS, a configuration that combines a shared backbone, a private residual adapter and classifier, geometric weights assigned by descending update norm, feature alignment, and private-parameter perturbations. The server computes a weighted sum of updates obtained from the same broadcast model; it does not obtain an additional optimization effect from sequential addition. A fully specified evaluation comprises 80 final runs: eight configurations, two datasets, and five training seeds on one fixed partition per dataset. On two-class-per-client CIFAR-10, ORDERS achieves 80.51 ± 0.79% native mean client accuracy, compared with 79.02 ± 1.42% for FedPer-R1 and 80.27 ± 0.73% for the matched uniform-weight control. After common local fine-tuning, the difference from FedPer-R1 narrows to 0.32 percentage points. On Sent140, ORDERS reaches 74.71 ± 0.49%, only 0.69 points above a post hoc client training-majority diagnostic. Ablations provide limited, endpoint-dependent evidence for norm ranking and alignment, and no clear benefit from perturbations. Parameter-payload savings are 5.47% and 0.78%, respectively. The study supplies auditable results and a precise account of the method, while showing that these records do not establish universal accuracy, fairness, communication-efficiency, or adversarial-robustness advantages.

Keywords: Personalized federated learning, Heterogeneous data, Weighted aggregation, Representation learning, Reproducible evaluation

## 1. Introduction

Federated learning coordinates local training while retaining training examples at clients. FedAvg combines local updates to produce a shared model [1]. When client distributions differ, the representation that benefits collaboration need not determine the best decision rule for each client. Personalized federated learning therefore permits client-specific predictors, including shared bodies with local prediction layers [2–4] and complete personal models coupled to a global model [5].

An additional design choice is how much influence each returned update receives. Sample-size or equal-client averaging supplies a reference; loss-based objectives, similarity-based collaboration, gradient-conflict modification, and learned aggregation offer alternatives [6–11]. These precedents mean that neither a shared/private split nor the combination of personalization and nonuniform aggregation is, by itself, a new principle. A useful empirical question is narrower: what does a particular weighting rule contribute when the architecture, local training procedure, and evaluation endpoint are made explicit?

ORDERS assigns decreasing geometric coefficients to shared-model updates ranked by descending Euclidean norm. The evaluated configuration also contains a private residual feature adapter, a local classifier, an alignment penalty, and occasional private-parameter perturbations. All participating clients begin their shared updates from the same broadcast state. Consequently, the aggregation operator is a normalized weighted sum. Its coefficients depend on rank, but permuting the additions of already weighted vectors creates no additional curvature or conflict-resolution mechanism.

This paper replaces earlier aggregate summaries with the recorded ORDERS-R1 evaluation. It reports eight configurations on CIFAR-10 and Sent140 over five final training seeds, including matched controls that remove rank weighting, alignment, or perturbations individually. We distinguish the native endpoint from a secondary endpoint with a common local fine-tuning budget. We also report client dispersion, lower-tail accuracy, parameter-payload accounting, and a simple local-majority diagnostic for the sentiment task. Predictions, run records, and an independent metric-recomputation script support the numerical tables. The contributions are threefold. First, we give an executable interpretation of the shared/private model and norm-rank operator, together with mathematical limits on claims about descent and reliability. Second, we provide a controlled, seed-paired empirical evaluation of the complete configuration and three component removals, with explicit baseline adaptations and tuning failures. Third, we show how endpoint choice, client label imbalance, and the actual parameter split change the interpretation of apparent personalization and communication gains. The evidence supports a modest CIFAR-10 benefit under the specified native evaluation, but much weaker conclusions for Sent140 and after common fine-tuning. No new architectural family, universal descent guarantee, or Byzantine defense is claimed.

## 2. Related work

### 2.1. Shared representations and personalized predictors

FedPer retains personalization layers locally while collaborating through base layers [2]. FedRep learns a shared representation and client heads using separate local optimization phases [3]. FedBABU trains the body with a fixed randomly initialized head during federation and adapts the head for personalization afterward [4]. These methods establish the architectural and training precedents for body/head separation. ORDERS-R1 retains a residual adapter as well as a classifier, but this implementation choice does not establish novelty for shared/private partitioning.

FedRoD separates generic and personalized prediction tasks, using a loss designed for heterogeneous label distributions for the generic component and local empirical risk for personalization [12]. FedProto communicates class prototypes and regularizes local training using aggregated prototypes [13]. Thus FedProto is closely related through representation collaboration, although its communication object differs from body parameters. FedProto's 2021 preprint and 2022 AAAI publication are both prior work; the published reference is by Tan et al. We discuss both methods as important untested comparators, rather than equating all collaborative/local designs with the parameter split studied here.

Ditto trains complete personal models with regularization toward a global model [5]. MOON uses model-level contrastive representation learning to correct local training under heterogeneity [14]; its original formulation primarily targets a global model, rather than the persistent private adapter used here. FedAMP learns personalized collaboration through attentive message passing among clients [6]. These works already connect local objectives with cross-client information exchange. ORDERS differs operationally by using one shared server body and norm-derived scalar coefficients, with squared within-predictor feature alignment instead of contrastive model comparisons or client-specific attentive model mixtures. This distinction is a method specification, not evidence of superiority.

### 2.2. Weighting, fairness, and robust aggregation

FedFV detects gradient conflicts using cosine similarity and modifies gradient directions and magnitudes before aggregation [7]. TERM changes the influence of losses through exponential tilting [8]; q-FFL emphasizes performance allocation through a loss-dependent objective [15]. Agnostic federated learning optimizes against mixtures of client distributions [9], while superquantile aggregation targets the upper tail of client losses [10]. These mechanisms should not all be characterized as equivalent rank heuristics. In particular, norm ranking does not rank losses, identify disadvantaged clients, or minimize a specified tail-risk objective. FedLAW additionally learns aggregation weights and a global shrinking factor [11], whereas ORDERS uses fixed normalized geometric coefficients.

FedCAP combines update calibration, customized aggregation, anomaly detection, and personalization to address heterogeneity and Byzantine attacks [16]. FLTrust uses a small clean server dataset to obtain a reference direction, assigns trust scores from directional agreement, and normalizes update magnitudes [17]. ORDERS uses neither a trusted reference nor a malicious-client detector; its descending norm score can favor a scaled malicious update. We therefore separate personalized predictive performance from adversarial robustness. The present campaign includes neither method as a trained baseline and contains no poisoning experiment.

Table 1. Mechanistic position of ORDERS. Literature comparisons are qualitative; only the R1 adaptations specified in Table 3 were run.

| Prior approach | Established mechanism | ORDERS-R1 distinction |
|---|---|---|
| FedPer, FedRep, FedBABU | Body/head separation; different local training and adaptation policies | Private residual adapter and head; norm-weighted body aggregation |
| FedRoD, FedProto | Decoupled prediction tasks or prototype exchange | One adapted feature path; shared parameter updates |
| MOON, FedAMP, Ditto | Contrastive correction, attentive collaboration, or personal-model regularization | Squared residual-feature penalty with one shared body |
| FedFV, TERM, q-FFL, AFL, superquantile FL | Conflict modification or explicit loss/distribution objectives | Displacement-norm ranking; no fairness objective |
| FedCAP, FLTrust | Calibrated personalized defense or trust-based normalized aggregation | No detection, clipping, or trusted reference |

## 3. ORDERS-R1 specification

### 3.1. Predictor and local objective

Client $i$ has training set $D_i$, $i = 1, \dots, N$. Let $w_t^G$ denote the shared body at the start of round $t$, and let $w_{i,t}^L = (A_{i,t}, V_{i,t}, b_{i,t})$ denote its retained private adapter and classifier. For a generic current local state, define

$$h = \phi^G(x; w^G), \quad \phi_i^L(x; w^G, w_i^L) = h + A_i h, \quad f_i(x) = V_i \phi_i^L(x) + b_i. \tag{1}$$

Both feature maps have dimension 64. The adapter $A_i \in \mathbb{R}^{64\times 64}$ has no bias and is initialized to zero. The classifier produces task logits. The local map depends on the shared body; there are not two independent encoders. We use "shared" and "private" to describe parameter ownership. Local retention alone does not give differential privacy or protection against inference from transmitted updates.

For minibatch $B \subset D_i$, the ORDERS training loss is

$$\mathcal{L}_i(B) = \frac{1}{|B|} \sum_{(x,y)\in B} \left[\ell_{\mathrm{CE}}(f_i(x), y) + \frac{\beta}{2} \|\phi^G(x) - \phi_i^L(x)\|_2^2\right], \quad \beta = 0.5. \tag{2}$$

The squared norm sums feature coordinates before averaging examples. Both feature paths receive gradients; neither is detached. Since their difference is $-A_i h$, the penalty is exactly $(\beta/2) \parallel A_i h \parallel^2$. It constrains the residual representation used for personalization and also affects the body through $h$. It can therefore discourage a useful local departure. The value 0.5 is fixed in this campaign, not selected by a sweep. The no-alignment control sets $\beta = 0$; it does not establish an optimal positive value or an adaptive per-client rule.

Each selected client performs two local passes using SGD with momentum 0.9 and weight decay $10^{-4}$. At the end of each pass, an independent Bernoulli event with probability 0.1 adds independent Gaussian noise of coordinate standard deviation 0.001 to every private tensor, including the adapter, classifier weight, and classifier bias. Shared tensors are not perturbed. Noise and minibatch ordering use separate seeded streams. These perturbations are a training component, not a privacy mechanism or demonstrated robustness defense.

Table 2. One synchronous ORDERS-R1 round. Unselected private states persist unchanged.

| Step | Operation |
|---|---|
| 1 | Sample $m_t$ clients without replacement and broadcast $w_t^G$. |
| 2 | Each selected client combines $w_t^G$ with its stored $w_{i,t}^L$. Reset its local SGD state and train for two passes with Eq. (2) and the specified private perturbations. |
| 3 | Retain the updated private tensors and return $\Delta_{i,t}$. |
| 4 | Compute shared-update norms; sort by decreasing norm and then client identifier. |
| 5 | Assign normalized geometric coefficients and apply Eq. (4). |

### 3.2. Server operator

For selected clients $S_t$, with $m_t = |S_t|$, let $\tilde{w}_{i,t}^G$ be the locally trained body and define

$$\Delta_{i,t} = \tilde{w}_{i,t}^G - w_t^G, \qquad s_{i,t} = \| \Delta_{i,t} \|_2. \tag{3}$$

The norm includes every shared-body coordinate. The server sorts descending norms, breaking ties by client identifier, to obtain $\pi_t$. With $0 < \omega \leq 1$,

$$\alpha_{j,t} = \frac{\omega^{j-1}}{\sum_{k=0}^{m_t-1} \omega^k}, \quad d_t = \sum_{j=1}^{m_t} \alpha_{j,t} \Delta_{\pi_t(j),t}, \quad w_{t+1}^G = w_t^G + \eta_s d_t. \tag{4}$$

The campaign fixes $\omega = 0.9$ and $\eta_s = 1$. At this server step size, the next body is a convex combination of returned bodies. At $\omega = 1$, it is equal-client averaging of the same updates. This differs from sample-size weighting when clients have different training-set sizes.

Norm computation and aggregation require $O(m_t d_G)$ work for $d_G$ shared coordinates, plus $O(m_t \log m_t)$ sorting. A norm reflects displacement, which also depends on the learning rate, local work, curvature, and parameterization. It does not identify update quality, alignment with a target objective, or client trustworthiness.

### 3.3. Schedule concentration and adversarial limitation

Even a fixed decay gives different concentration as participation changes. The top-to-bottom coefficient ratio is $\omega^{-(m_t-1)}$, and an effective number of weighted contributors is $n_{\text{eff}} = 1/\sum_j \alpha_{j,t}^2$. With $\omega = 0.9$, the ten-client CIFAR-10 rounds have top/bottom ratio 2.58 and $n_{\text{eff}} = 9.18$; the twenty-client Sent140 rounds have ratio 7.40 and $n_{\text{eff}} = 14.88$. All coefficients remain positive, but lower-ranked legitimate clients can receive substantially less influence. These are algebraic properties of the schedule, not measured advantages or evidence that it is optimal. No automated schedule-selection mechanism is implemented.

A client allowed to submit an arbitrary vector can increase its norm to obtain a top rank. Once largest, an update $Mv$ for unit vector $v$ contributes $\alpha_{1,t} Mv$, which is unbounded as $M$ grows when the other updates remain fixed. Its scalar coefficient is 1.54 times the uniform coefficient for ten participants and 2.28 times for twenty. These coefficient ratios are not attack-success measurements; they expose why norm ranking alone cannot control priority hijacking. The evaluated method does not clip updates or reject suspicious clients. A bounded or trust-based alternative would be a different algorithm requiring adaptive-attack evaluation.

## 4. Mathematical interpretation and limits

### 4.1. Weighted addition and a valid descent statement

For fixed updates and coefficients, an accumulator $z_0 = w_t^G$, $z_k = z_{k-1} + \eta_s q_{\sigma(k)} \Delta_{\sigma(k),t}$ ends at Eq. (4) for every permutation $\sigma$, where $q_{\pi_t(j)} = \alpha_{j,t}$. This follows by telescoping. Reassigning coefficients can change the aggregate; rearranging the same weighted additions cannot, apart from floating-point rounding. The client gradients are not re-evaluated at intermediate accumulator states.

The following standard smoothness statement describes what can be asserted for a fixed differentiable objective. It is not a claim that the personalized training process minimizes that objective.

**Proposition 1 (Conditional comparison of aggregate steps).** Let $F$ have an $L$-Lipschitz gradient on a convex region containing the relevant segments, let $d$ and $u$ be fixed directions, and let $\eta_s > 0$. Then

$$F(w+\eta_s d) - F(w) \le \eta_s \langle \nabla F(w), d\rangle + \frac{L\eta_s^2}{2} \parallel d \parallel^2, \tag{5}$$

$$F(w+\eta_s d) - F(w+\eta_s u) \le \eta_s \langle \nabla F(w+\eta_s u), d-u\rangle + \frac{L\eta_s^2}{2} \parallel d-u \parallel^2. \tag{6}$$

Thus a strictly negative right-hand side of Eq. (6) is sufficient for a strictly lower objective value from d than from u.

The proof appears in Appendix A. With $u = m_t^{-1} \sum_i \Delta_{i,t}$, the condition involves the direction of $d-u$ relative to the objective gradient, not just pairwise update conflicts. Neither that gradient nor a certified smoothness constant was measured in the campaign, so this condition supplies no numerical prediction of the observed accuracy differences. Accuracy differences are not objective-descent estimates.

### 4.2. Why conflict does not imply an advantage

Consider $F(w) = \frac{1}{2} \parallel w \parallel^2$ and two updates $\Delta_+ = -w + v$, $\Delta_- = -w - v$ with $\parallel v \parallel > \parallel w \parallel$. Their inner product is negative. Uniform averaging with $\eta_s = 1$ reaches zero, the exact minimizer. Assigning coefficients $\theta$ and $1-\theta$ instead yields

$$w^+ = (2\theta - 1)v, \quad F(w^+) - F(0) = 1/2\,(2\theta-1)^2 \parallel v \parallel^2 > 0 \quad (\theta \neq 1/2). \tag{7}$$

This refutes universal superiority of nonuniform weighting in conflicting-update settings. There need be no norm tie: in one dimension, $w = \varepsilon$ and $v = 2\varepsilon$ give updates $\varepsilon$ and $-3\varepsilon$. Descending-norm geometric weighting incurs loss $2[(1-\omega)/(1+\omega)]^2\varepsilon^2$, while uniform averaging attains zero. The example persists for arbitrarily small $\varepsilon$.

The earlier curvature–conflict claim is therefore withdrawn, rather than repaired by a sign change. A term $-cC$ with $c > 0$ and $C < 0$ is positive and cannot be interpreted as additional descent in an upper bound on loss change. Furthermore, positive semidefiniteness does not permit a signed cross-term bound $x^\top H y \le \lambda_{\min} x^\top y$. Appendix A gives an explicit example and a generic bias-aware inequality. We claim no unconditional convergence rate for ORDERS: norm-derived coefficients depend on the updates, and the private states evolve, so unbiased mean-objective gradients do not follow from unbiased individual client estimators.

## 5. Experimental protocol

### 5.1. Data and partitions

The recorded protocol is ORDERS-R1-2026-09-29-v1. All methods use the same data partition, generated with seed 1729, and final training seeds 101, 202, 303, 404, and 505. These seeds vary training randomness, not the partition. The study contains 80 final training runs (8 × 2 × 5), each with 100 rounds; every run has two evaluation endpoints.

CIFAR-10 [18] is divided among 50 clients, each with exactly two classes. Five random label cycles allocate balanced class ownership. The official training/test boundary is preserved, and a stratified 10% of each client's official training examples is used for validation. Each client has 900 training, 100 validation, and 200 test examples, giving totals of 45,000, 5,000, and 10,000. Images use fixed channel normalization without random augmentation.

Sent140 uses the original source training tweets and user identifiers, following the user-client motivation of LEAF [19], but it uses a fresh split rather than the historical LEAF split. A seeded random order selects eligible users; exact tweet identifiers and normalized text are deduplicated globally over candidate users before final eligibility checks. There are 100 retained users with at least 50 deduplicated tweets and a cap of 1,000 per user. The candidate deduplication removes 420 records. Within-user stratified splitting yields 5,758 training, 1,245 validation, and 1,245 test examples from 8,248 tweets. Test sets contain 7–51 examples per client. A vocabulary capped at 8,000 entries is learned from training examples only; sequences are truncated to 40 tokens and use no pretrained embeddings. Example identifiers are disjoint across the recorded clients and splits. Appendix B details preprocessing and source hashes.

### 5.2. Models and optimization

All methods use the same complete predictor form in Eq. (1). The CIFAR body has four 3 × 3 convolutions with channels 32, 32, 64, and 64, ReLU activations, and max-pooling after each pair. Adaptive 2 × 2 pooling and a shared 256 → 64 projection produce the feature vector. Sent140 uses a 64-dimensional embedding and a one-layer LSTM with hidden dimension 64; packed sequences supply the final hidden state. Neither model uses batch normalization or dropout. In split methods, the body is shared and the adapter and classifier remain local. The total float32 parameter counts are 86,762 for CIFAR-10 and 541,634 for Sent140, of which 82,016 and 537,408 are shared, respectively.

Each round samples 20% of clients without replacement: ten for CIFAR-10 and twenty for Sent140. Selection and initialization are paired across methods within a seed. The local budget is two full passes through participating clients' training data. SGD uses batch size 64, momentum 0.9, and weight decay $10^{-4}$; optimizer state is reset at each local phase and round. Table 3 specifies how each method allocates those passes. Equal data passes do not imply equal FLOPs, trainable-parameter work, memory, or final adaptation cost.

Ditto-R1 regularizes all personal parameters toward the pre-round broadcast global model with $(\lambda/2)\sum_k \| w_{i,k} - w_{t,k} \|^2$, where $\lambda = 0.1$. Its global and personal passes partition the two-pass budget. FedRep-R1 uses only one private pass per body pass, rather than exploring larger head-training budgets. These choices, the shared residual-adapter architecture, and the limited tuning grid constrain comparisons with the published methods.

### 5.3. Tuning and evaluation endpoints

Each of the five complete methods is tuned over learning rates {0.01, 0.03, 0.1} using 20 rounds and seed 9001. Selection uses final-round native mean client validation accuracy, with a smaller-learning-rate tie break among finite candidates. All complete methods select 0.01 on CIFAR-10 and 0.1 on Sent140. The three ORDERS controls inherit the selected ORDERS learning rate so that they measure fixed-configuration removals. Other hyperparameters are fixed; they are not described as optimal. Four CIFAR-10 candidates at learning rate 0.1 stop with nonfinite losses (FedAvg-FT, FedPer, Ditto, and ORDERS); 26 candidates finish. All failures and finite results are retained in Appendix C. All 80 final runs finish. Final tests use round 100 after tuning is locked, with no test-based checkpoint selection.

At the native endpoint, split methods combine the final server body with each client's stored private tensors. Ditto uses its persistent complete personal model. FedAvg-FT uses two train-only, full-model fine-tuning passes from the final global state. Thus native performance includes each method's stated personalization policy. At the common FT endpoint, two task-loss-only full-model passes are applied to each corresponding pre-adaptation predictor. FedAvg-FT reuses its already fine-tuned state and is not adapted twice. The adaptation learning rate remains the selected method/dataset rate; momentum is reset. Common FT equalizes the additional adaptation budget and loss, but does not erase differences in preceding local training.

Table 3. Implemented R1 configurations. “Private” includes both adapter and classifier. All baseline labels retain R1 because these are specified adaptations, not reproductions of the original publications’ benchmark settings.

| Method | Local procedure per participation | Server aggregation |
|---|---|---|
| FedAvg-FT-R1 | Two full-model passes; two final train-only full-model fine-tuning passes per client | Sample-weighted complete models |
| FedPer-R1 | Two joint body/private passes | Sample-weighted bodies |
| FedRep-R1 | One private pass with body frozen, then one body pass with private parameters frozen | Equal-client bodies |
| Ditto-R1 | One global-model pass; one persistent personal-model pass with parameter penalty | Sample-weighted complete global models |
| ORDERS-R1 | Two joint passes, $\beta = 0.5$ alignment, private perturbations | Norm-rank bodies, $\omega = 0.9$ |
| ORDERS-uniform-R1 | Same as ORDERS | Equal-client bodies |
| ORDERS-no-align-R1 | Same as ORDERS except $\beta = 0$ | Norm-rank bodies |
| ORDERS-no-mutation-R1 | Same as ORDERS except perturbation probability zero | Norm-rank bodies |

### 5.4. Metrics and statistical scope

For run $r$, let $a_{i,r}$ be client test accuracy in percent. The primary score is the equal-client mean $A_r = N^{-1} \sum_i a_{i,r}$. Pooled accuracy weights clients by test size and is reported separately in the evidence files. We summarize $A_r$ by its five-seed mean and sample standard deviation (SD, denominator four). Pairing uses corresponding training seeds on the same split. For ORDERS-minus-comparator differences $d_r$, descriptive 95% intervals are

$$\overline{d} \pm t_{4,0.975}\, s_d/\sqrt{5}. \quad (8)$$

These intervals are unadjusted, have only five paired observations, and describe training variability conditional on one partition. They do not cover partition or client-population uncertainty. Clients are not treated as independent training repetitions.

For transparency, Appendix D also reports exploratory paired t-test probabilities with a Holm correction across all 28 ORDERS-minus-comparator accuracy comparisons (7 comparators, 2 datasets, 2 endpoints). This family was chosen during analysis, not preregistered. The correction is a sensitivity check, not a basis for selecting favorable comparisons. We emphasize effect sizes, endpoint dependence, and uncertainty rather than declaring a universal winner.

We measure client disparity using the population SD within each run, $[N^{-1} \sum_i (a_{i,r} - A_r)^2]^{1/2}$, and the mean accuracy of the lowest $\lceil 0.1N \rceil$ clients, then average these measures over seeds. They describe performance dispersion and lower-tail outcomes, not demographic fairness or a constrained optimization guarantee. Communication is accounted for as transmitted parameter bytes, as specified in Section 6.4.

## 6. Results

### 6.1. Accuracy and endpoint dependence

Table 4 contains every final configuration and both endpoints; Figure 1 shows the complete-method comparisons. At the native CIFAR-10 endpoint, ORDERS attains 80.51 ± 0.79%. Its paired difference from FedPer-R1, the strongest native baseline in this set, is 1.494 pp with unadjusted 95% interval [0.294, 2.694]. Differences from FedAvg-FT-R1, FedRep-R1, and Ditto-R1 are 3.616, 4.188, and 7.404 pp, respectively. These are comparisons between complete R1 procedures. They do not attribute all differences to rank weighting.

Common fine-tuning raises the CIFAR-10 means for ORDERS and FedPer-R1 to 81.16% and 80.84%. Their paired gap narrows to 0.318 pp, with interval [−0.142, 0.778]. Consequently, the native gap from FedPer should not be described as a stable advantage under all personalization policies. FedRep-R1 and Ditto-R1 remain lower in this campaign, but their restricted local phase budgets and hyperparameter settings limit extrapolation to optimally tuned implementations.

Sent140 provides weaker differentiation. ORDERS native accuracy is 74.71 ± 0.49%, compared with 74.64 ± 0.33% for FedRep-R1 and 74.55 ± 0.30% for FedPer-R1. The paired ORDERS-minus-FedRep difference is 0.065 pp with interval [−0.449, 0.580]. Under common FT, ORDERS falls to 74.35%, whereas FedRep-R1 reaches 74.73%; their difference is −0.375 pp with interval [−1.207, 0.458]. All Sent140 common-FT comparison intervals include zero. More local training does not uniformly improve this endpoint.

Across the exploratory family of 28 paired comparisons, only CIFAR-10 comparisons with FedAvg-FT-R1, FedRep-R1, and Ditto-R1 retain Holm-adjusted $p < 0.05$ at both endpoints. Comparisons with FedPer-R1, all component comparisons, and all Sent140 comparisons do not. This does not establish equivalence for the latter comparisons; it limits the strength of superiority claims from this small campaign.

Table 4. Final mean client test accuracy (%), mean ± sample SD across five training seeds. Both datasets use one fixed partition. All rows are R1 configurations; the suffix is omitted here for space. The two endpoint definitions are in Section 5.3.

| Method | CIFAR-10 Native | CIFAR-10 Common FT | Sent140 Native | Sent140 Common FT |
|---|---|---|---|---|
| FedAvg-FT | 76.90 ± 0.48 | 76.90 ± 0.48 | 73.84 ± 0.43 | 73.84 ± 0.43 |
| FedPer | 79.02 ± 1.42 | 80.84 ± 0.64 | 74.55 ± 0.30 | 74.35 ± 0.28 |
| FedRep | 76.33 ± 0.98 | 75.65 ± 1.67 | 74.64 ± 0.33 | 74.73 ± 0.38 |
| Ditto | 73.11 ± 1.25 | 73.82 ± 1.16 | 74.19 ± 0.20 | 74.33 ± 0.32 |
| ORDERS | 80.51 ± 0.79 | 81.16 ± 0.67 | 74.71 ± 0.49 | 74.35 ± 0.56 |
| ORDERS: uniform | 80.27 ± 0.73 | 80.80 ± 0.57 | 74.57 ± 0.41 | 74.19 ± 0.55 |
| ORDERS: no alignment | 78.91 ± 1.59 | 80.91 ± 0.83 | 74.62 ± 0.45 | 74.21 ± 0.68 |
| ORDERS: no mutation | 80.60 ± 0.81 | 81.10 ± 0.66 | 74.70 ± 0.53 | 74.33 ± 0.58 |

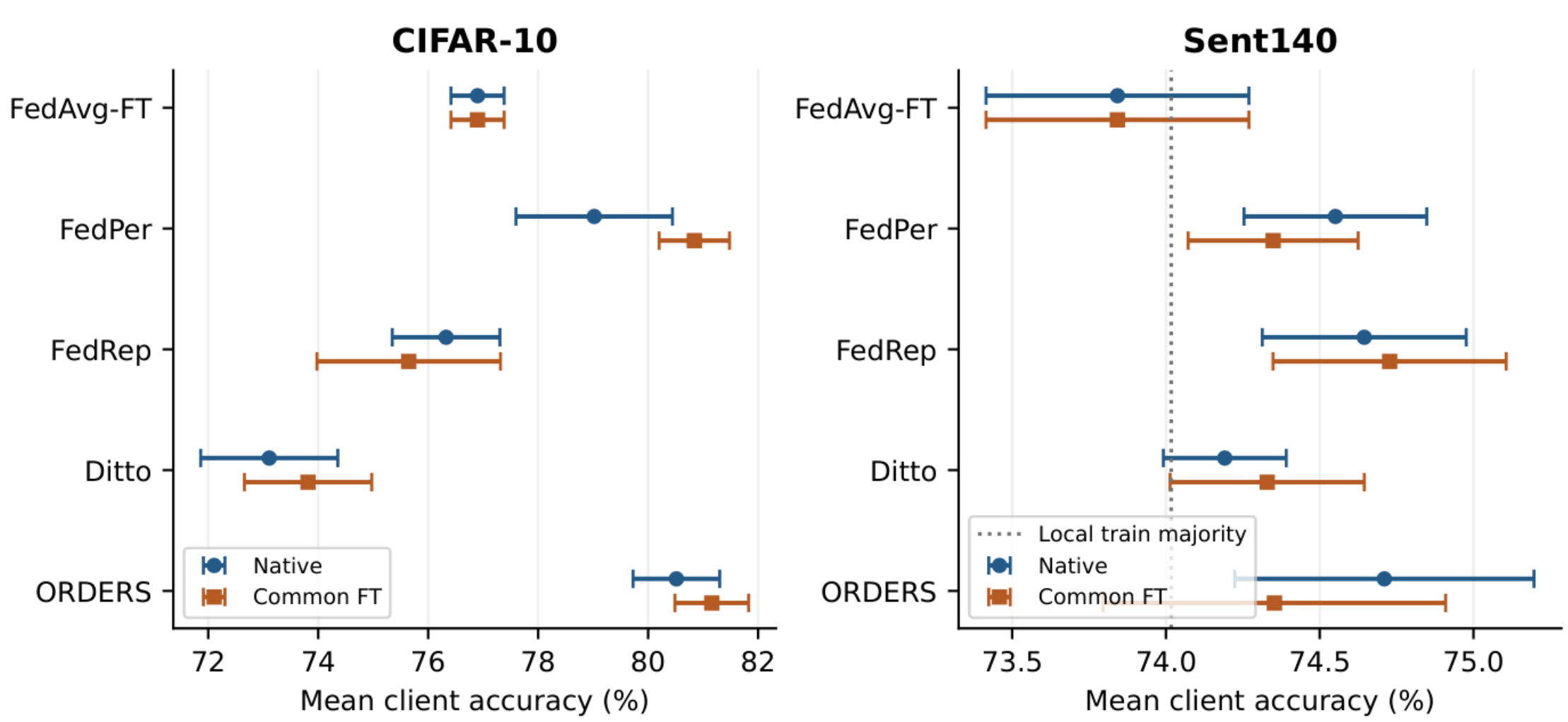

Figure 1. Complete-method accuracy at the two endpoints. Bars indicate one sample SD across five training seeds, not confidence intervals. The dotted Sent140 line is the post hoc per-client training-majority predictor evaluated on test data. Dataset panels use different horizontal scales.

## 6.2. Component ablations

Table 5 gives paired effects for each single-component removal; Table 4 gives their absolute accuracies. Positive effects favor the complete configuration. On native CIFAR-10, replacing uniform weights by the norm-rank rule gives a mean difference of 0.246 pp, with interval [0.062, 0.430]. Removing alignment yields a larger but less precise gap of 1.606 pp, with interval [0.115, 3.097]. Both unadjusted intervals exclude zero, but neither comparison survives the exploratory correction. After common FT, these differences become 0.354 pp and 0.244 pp, with intervals including zero.

Private perturbations have no clearly established benefit. On native CIFAR-10, the no-mutation control reaches 80.60%, slightly above ORDERS at 80.51%; the paired complete-minus-control difference is −0.086 pp with interval [−0.235, 0.063]. Sent140 component effects are all small and all intervals include zero at both endpoints. The records therefore do not justify presenting all three components as necessary for improved performance. The controls hold the ORDERS learning rate fixed; they estimate removal effects at that configuration, not each variant's best attainable performance. Interactions among removed components are not tested.

Table 5. Ablation effects: ORDERS minus the named control, in pp. Intervals are unadjusted paired 95% t intervals across five seeds. The final column applies the exploratory 28-comparison Holm correction to paired tests. All rows use the corresponding R1 configurations.

| Endpoint | Control | Mean effect | 95% interval | Holm p |
|---|---|---|---|---|
| **CIFAR-10** | | | | |
| Native | Uniform | 0.246 | [0.062, 0.430] | 0.453 |
| Native | No alignment | 1.606 | [0.115, 3.097] | 0.765 |
| Native | No mutation | -0.086 | [-0.235, 0.063] | 1.000 |
| Common FT | Uniform | 0.354 | [-0.268, 0.976] | 1.000 |
| Common FT | No alignment | 0.244 | [-0.330, 0.818] | 1.000 |
| Common FT | No mutation | 0.062 | [-0.032, 0.156] | 1.000 |
| **Sent140** | | | | |
| Native | Uniform | 0.139 | [-0.122, 0.399] | 1.000 |
| Native | No alignment | 0.088 | [-0.186, 0.362] | 1.000 |
| Native | No mutation | 0.015 | [-0.105, 0.135] | 1.000 |
| Common FT | Uniform | 0.166 | [-0.232, 0.564] | 1.000 |
| Common FT | No alignment | 0.139 | [-0.207, 0.486] | 1.000 |
| Common FT | No mutation | 0.022 | [-0.039, 0.084] | 1.000 |

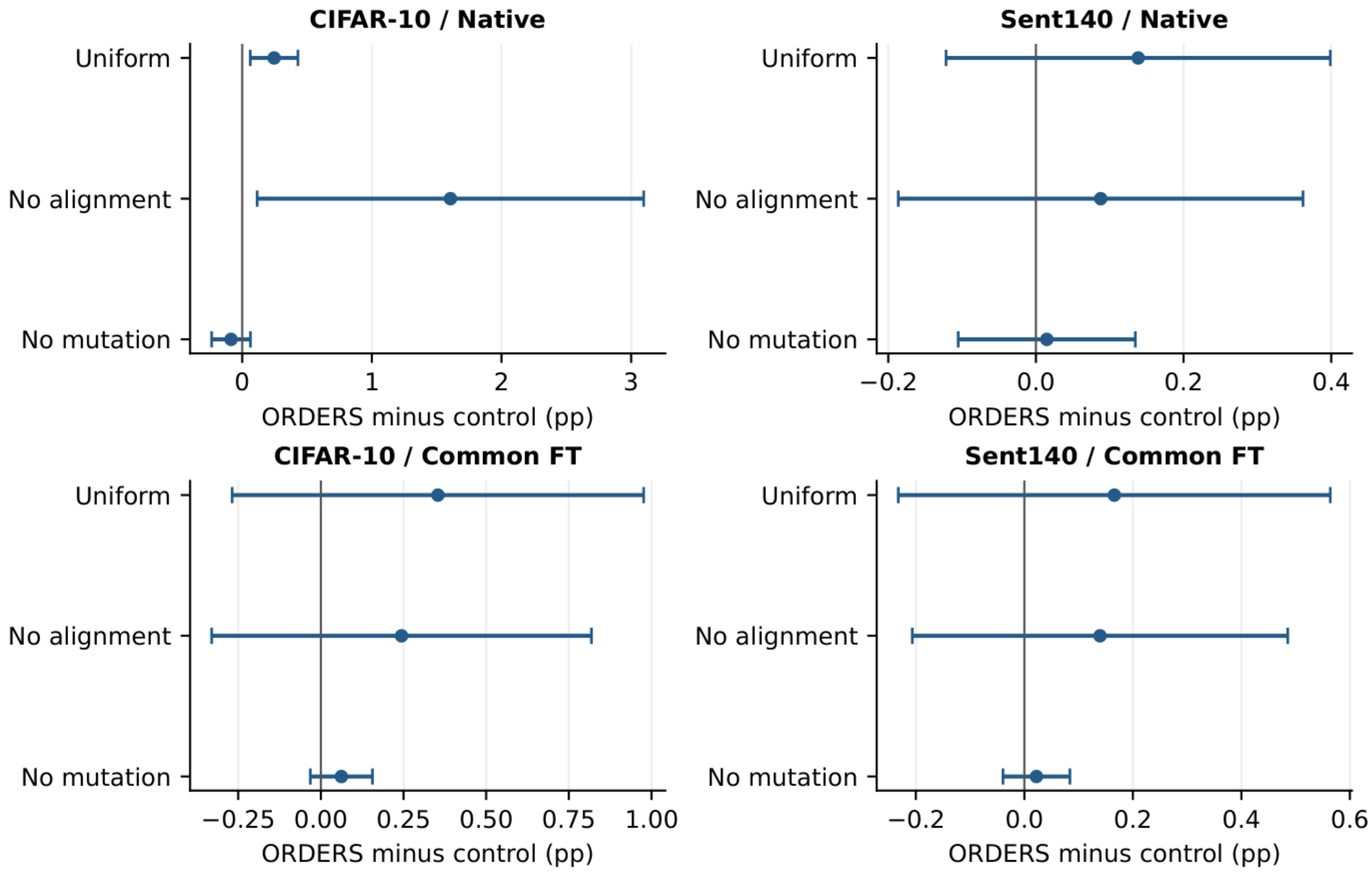

Figure 2. Paired component effects from Table 5. Bars are unadjusted 95% intervals. Positive values favor complete ORDERS. Horizontal scales vary by panel to make the uncertainty visible; none of these comparisons has Holm-adjusted $p < 0.05$ in the exploratory family.

### 6.3. Client outcomes and the Sent140 majority diagnostic

Table 6 reports native client dispersion and bottom-decile accuracy. On CIFAR-10, ORDERS has lower mean client SD than the four complete baselines and a bottom-decile mean of 57.54%, compared with 56.54% for FedPer-R1 and 57.32% for uniform ORDERS. The paired bottom-decile differences are uncertain: 1.00 pp versus FedPer with interval [−1.10, 3.10], and 0.22 pp versus uniform with interval [−0.76, 1.20]. The no-mutation control is slightly better on both displayed CIFAR-10 measures. On Sent140, FedRep-R1 has lower dispersion and higher bottom-decile accuracy than ORDERS. A universal fairness advantage is not supported.

To assess the role of user label imbalance, we constructed a post hoc diagnostic that always predicts each client's most frequent training label, breaking ties toward the smaller label. This uses no validation or test labels to select the prediction. Its Sent140 mean client accuracy is 74.017%, while its pooled accuracy is 75.984%; the difference illustrates why these averaging conventions cannot be interchanged. ORDERS native mean client accuracy exceeds this diagnostic by only 0.693 pp. Across its five native runs, 61–67 of the 100 clients receive constant test predictions, and 90.20–91.89% of individual predictions agree with the client's training-majority label. These counts do not prove that representations contain no useful information, but they substantially limit claims of broad language-modeling improvement from the mean accuracy alone. The same local-majority diagnostic is 50% on the balanced two-class CIFAR-10 clients. This diagnostic was not part of tuning and is not an additional stochastic training baseline.

Table 6. Native client-level outcomes, averaged over five seeds. Client SD uses population SD across clients within a run (lower is less dispersed); bottom 10% is the mean accuracy of the lowest five CIFAR-10 or ten Sent140 clients (higher is better). These are distinct from the seed SDs in Table 4.

| Method (R1) | CIFAR-10 Client SD (pp) | CIFAR-10 Bottom 10% (%) | Sent140 Client SD (pp) | Sent140 Bottom 10% (%) |
|---|---|---|---|---|
| FedAvg-FT | 13.80 | 53.04 | 17.16 | 41.07 |
| FedPer | 12.98 | 56.54 | 16.84 | 42.03 |
| FedRep | 13.67 | 53.78 | 16.36 | 42.91 |
| Ditto | 12.72 | 51.28 | 16.82 | 41.39 |
| ORDERS | 12.34 | 57.54 | 16.68 | 42.36 |
| ORDERS: uniform | 12.47 | 57.32 | 16.63 | 42.27 |
| ORDERS: no alignment | 13.12 | 55.60 | 16.70 | 42.53 |
| ORDERS: no mutation | 12.32 | 57.72 | 16.70 | 42.32 |

## 6.4. Communication accounting

Let $B_G$ be shared-body bytes, $B_L$ private bytes, and $B = B_G + B_L$. A participating split-method client uploads $B_G$ bytes and downloads $B_G$ bytes per training round. For $T$ rounds with $m$ participants each, the accounted payload is $2TmB_G$, compared with $2TmB$ for the full-model reference. Their fixed-budget ratio is

$$r_{\text{payload}} = \frac{B_G}{B}, \qquad \text{saving} = 100(1 - r_{\text{payload}})\%. \tag{9}$$

Table 7 uses the actual float32 tensor inventories. Savings are 5.47% for CIFAR-10 and 0.78% for Sent140, because most parameters belong to the shared body. A two-part architecture does not imply that each part contains half the parameters.

The accounting excludes initialization, protocol messages, serialization, evaluation, and network headers. It is neither measured bandwidth nor measured latency. With equal rounds and participation, the per-round parameter ratio also equals the ratio over that fixed training horizon; no faster-convergence assumption is needed for this identity. Communication to a common target accuracy additionally depends on the required number of rounds and is not evaluated here. Final adaptation is local, and its compute is separate from the training-round payload. There is no 0.5× communication claim in this study.

Table 7. Parameter-payload accounting per seed over 100 training rounds, both directions. Full-model reference is FedAvg-FT-R1; FedPer-R1 and FedRep-R1 share ORDERS' body-only payload. One MiB is $2^{20}$ bytes.

| Dataset | $B_G$ (bytes) | $B$ (bytes) | Ratio | Saving (%) | Split (MiB) | Full (MiB) |
|---|---|---|---|---|---|---|
| CIFAR-10 | 328,064 | 347,048 | 0.9453 | 5.47 | 625.73 | 661.94 |
| Sent140 | 2,149,632 | 2,166,536 | 0.9922 | 0.78 | 8200.20 | 8264.68 |

## 7. Discussion and limitations

The most defensible positive finding is the performance of the full configuration on the specified native CIFAR-10 task. The matched uniform control suggests that norm weighting has a small effect there, while alignment has a larger but less precise native removal effect. The reduced differences after common fine-tuning show why the final adaptation policy is part of the comparison. A gain over a baseline with a different local phase allocation is not a causal estimate of a server rule's value.

The Sent140 results demonstrate a different limitation: personalized label priors can explain much of a high mean local accuracy. The majority diagnostic, small test sets, and limited differences among complete methods counsel against interpreting the two datasets as universal evidence across modalities. This study contains image and sentiment tasks only; it includes no speech evaluation.

Architecture and hyperparameter sensitivity remain unresolved. We evaluate one body/adapter/classifier split in a small CNN and an LSTM. No layer-selection mechanism, Transformer, GNN, split sweep, decay sweep, or positive-β sweep is present. Alignment explicitly limits local residual features; whether a client benefits from a larger departure should be assessed with training/validation data in a separate study. Similarly, the fixed schedule's concentration grows with participant count. An adaptive weighting rule would need an objective, a selection protocol, and comparisons against learned or loss-aware aggregation; it cannot be inferred from this campaign.

The baseline set is incomplete. FedRoD, FedProto, FedBABU, FedAMP, MOON, FedFV, TERM, q-FFL, agnostic FL, superquantile FL, FedCAP, FLTrust, and FedLAW are discussed but not trained here. The present results establish no ordering relative to them. The R1 adaptations share an adapter architecture and a two-pass local budget; this supports an explicit comparison while differing from the original methods' preferred architectures and tuning practices. Only learning rate is tuned, using one short tuning seed. Instability at the largest CIFAR-10 rate illustrates sensitivity rather than robustness.

Five training seeds on one partition support only limited uncertainty estimates. The study does not assess alternative client populations, population shifts, dropout, variable local work, adversarial updates, or label-noise attacks. No global-only prediction endpoint was saved, so a global-to-local personalization gain cannot be computed. Private retention should not be confused with a formal privacy guarantee, nor norm weighting with robust aggregation. The observed fairness measures are descriptive outcomes rather than constraints enforced by the method.

Finally, the evidence audit establishes consistency of saved records and recomputation from saved predictions. It does not independently reacquire raw data or replay inference from model checkpoints. The exported records omit checkpoint tensors, although their hashes are indexed. These boundaries are explicit in the reproducibility statement and do not justify stronger claims about the unobserved training process.

## 8. Conclusion

ORDERS-R1 is a precisely specified combination of norm-rank-weighted body aggregation, persistent private adapters and heads, residual-feature alignment, and private perturbations. An 80-run evaluation finds a modest native CIFAR-10 advantage over the strongest included baseline, but a smaller difference after common fine-tuning and weak separation on Sent140. The measured rank-weighting effect is small, the alignment effect depends on the endpoint, and perturbations have no clear benefit. Actual shared-parameter sizes yield small payload savings. The contribution is an auditable empirical characterization of this configuration and its limitations, rather than a new shared/private architecture or a guarantee of conflict mitigation, fairness, or adversarial robustness.

## Data and code availability

The evidence package described in this study contains manuscript sources, verified CSV tables, the metric-audit script, the figure/table generator, and Evidence_Records.zip. The latter contains configurations, environment metadata, dataset manifests and split assignments, captured source, round logs, and saved predictions. Raw images and tweets are not redistributed in that archive. Dataset source locations and fingerprints appear in Appendix B. The records exclude 860 model files totaling 3,325,439,029 bytes, so the package supports saved-prediction metric verification but not checkpoint inference replay. No public repository or permanent data DOI is asserted. Appendix F states the completed checks.

## Declaration of generative AI and AI-assisted technologies in the manuscript preparation process

None.

## Appendix A. Mathematical details

### Appendix A.1. Smoothness and comparison proof

For a vector $v$ whose segment lies in the stated smoothness region,

$$F(w+v)-F(w)=\int_0^1 \langle \nabla F(w+sv), v\rangle\, ds$$

$$\leq \langle \nabla F(w), v\rangle + \int_0^1 L\, s \parallel v \parallel^2 \; ds = \langle \nabla F(w), v\rangle + \frac{L}{2} \parallel v \parallel^2 .$$

Substituting $v=\eta_s d$ proves Eq. (5). Applying the same result from base point $w+\eta_s u$ with displacement $\eta_s(d-u)$ proves Eq. (6). A negative upper bound in the latter establishes the stated strict comparison. A comparison of two unrelated upper bounds, by itself, would not establish an ordering between their left-hand sides.

For $d=\sum_i q_i\, \Delta_i$, its quadratic form can contain pairwise inner products through

$$\parallel \sum_i q_i\, \Delta_i \parallel^2 = \sum_i q_i^2 \parallel \Delta_i \parallel^2 + 2\sum_{i<j} q_i\, q_j \langle \Delta_i, \Delta_j\rangle .$$

This identity is part of the ordinary smoothness bound; it is not an extra sequential-path term. Nor does its sign determine the first-order term involving $\nabla F$. For the positive definite matrix $H=\mathrm{diag}(1{,}3)$, vectors $x=(1{,}1)^\top$, $y=(-2{,}1)^\top$ satisfy $x^\top y=-1$ and $x^\top H y=1$. Thus neither $\lambda_{\min} x^\top y$ nor $\lambda_{\max} x^\top y$ gives the asserted signed upper bound on a cross term. The valid bound $|x^\top H y| \leq \parallel H \parallel_{\mathrm{op}} \parallel x \parallel \parallel y \parallel$ does not preserve the sign of the Euclidean inner product.

### Appendix A.2. The quadratic counterexample and the withdrawn inequality

The scalar construction in Section 4.2 can be realized by one exact gradient step of size one on client losses $f_+(z)=\frac{1}{2}(z-v)^2$ and $f_-(z)=\frac{1}{2}(z+v)^2$. Their mean is $\frac{1}{2}z^2+\frac{1}{2}v^2$, differing from $F(z)=\frac{1}{2}z^2$ only by a constant. Uniform averaging therefore reaches the minimizer of the mean objective. With $w=\varepsilon$, $v=2\varepsilon$, descending norm assigns $1/(1+\omega)$ to $-3\varepsilon$ and $\omega/(1+\omega)$ to $\varepsilon$. The new state is $-2\varepsilon(1-\omega)/(1+\omega)$, giving the loss stated in the main text.

The earlier inequality also fails under uniform aggregation. With $m=2$, $\eta_s=\lambda_{\min}=1$, uniform coefficients, and conflict sum $C=\frac{1}{4}(\varepsilon)(-3\varepsilon)=-3\varepsilon^2/4$, its proposed right-hand side was

$$-1/2\, |\varepsilon-3\varepsilon|^2 - 1/2\, C = -13\varepsilon^2/8.$$

The exact loss change is $-\varepsilon^2/2$, and the quadratic has zero third-order remainder. The proposed upper bound is false. This calculation is a correction to the previous claim, not a new benchmark experiment. Equation (5) is exact for the uniform step in this example.

### Appendix A.3. What a bias-aware inequality would require

For completeness, fix private reference states $\overline{w}_i^L$ and a lower-bounded, $L$-smooth objective $F(w)=N^{-1}\sum_i F_i\,(w,\overline{w}_i^L)$. Consider a hypothetical recursion $w_{t+1}=w_t-\eta h_t$, with deterministic $w_0$, $0<\eta\leq 1/L$, finite relevant expectations, and history $\mathcal{H}_t$. Define

$$\mu_t = \mathbb{E}[h_t \mid \mathcal{H}_t], \quad b_t = \mu_t - \nabla F(w_t), \quad v_t = \mathbb{E}[\| h_t - \mu_t \|^2 \mid \mathcal{H}_t].$$

Smoothness and conditional expectation give

$$\mathbb{E}[F(w_{t+1}) \mid \mathcal{H}_t] \leq F(w_t) - \eta\langle\nabla F(w_t), \mu_t\rangle + \frac{L\eta^2}{2}(\| \mu_t \|^2 + v_t)$$

$$\leq F(w_t) - \frac{\eta}{2} \| \nabla F(w_t) \|^2 + \frac{\eta}{2} \| b_t \|^2 + \frac{L\eta^2}{2} v_t.$$

The second line uses $-2\langle g, \mu\rangle = -\| g \|^2 - \| \mu \|^2 + \| g - \mu \|^2$ and drops the nonpositive coefficient of $\| \mu_t \|^2$. Taking expectations, summing, and using $F \geq F_{\inf}$ yields

$$\frac{1}{T}\sum_{t<T} \mathbb{E} \| \nabla F(w_t) \|^2 \leq \frac{2(F(w_0) - F_{\inf})}{\eta T} + \frac{1}{T}\sum_{t<T} \mathbb{E} \| b_t \|^2 + \frac{L\eta}{T}\sum_{t<T} \mathbb{E}\, v_t. \quad \text{(A.1)}$$

This generic inequality retains the bias explicitly. Norm-dependent weights do not make $b_t$ vanish, and the evolving private states in ORDERS do not satisfy the fixed-reference setup automatically. No bounds on these terms are established by the experiments. Equation (A.1) is therefore not an ORDERS convergence rate. Sorting update norms also provides no identity with a loss-ranked spectral-risk gradient.

## Appendix B. Implementation and provenance

### Appendix B.1. Preprocessing and initialization

CIFAR-10 inputs are converted to float32, divided by 255, and normalized by channel means (0.4914, 0.4822, 0.4465) and standard deviations (0.2470, 0.2435, 0.2616). Convolutions use stride one, padding one, and biases. Max-pooling is 2 × 2. The final body projection has a bias and ReLU. The private classifier is linear with a bias. The only zero-initialized learned block is the residual adapter; remaining modules use PyTorch defaults under the paired seed.

The Sent140 loader reads training.1600000.processed.noemoticon.csv in Latin-1 and maps source labels 0 and 4 to binary labels 0 and 1. Candidate users are the first 400 in a seeded permutation of users with at least 50 source records; the first 100 retaining at least 50 records after deduplication are used. Deduplication compares original tweet IDs and lowercased, whitespace-normalized text. Tokenization lowercases text, replaces URLs with urltoken, and user mentions with usertoken; tokens comprise words with optional internal apostrophes, digit runs, and individual non-word non-whitespace characters. Vocabulary ties use lexical order after descending training frequency. Padding is index 0, unknown tokens index 1, and an empty sequence becomes one unknown token. The LSTM uses the final hidden state of the packed sequence. No raw tweet text is included in the evidence archive.

Within-client class lists are shuffled using deterministic derived seeds and split using the recorded stratified routine; the exact resulting memberships are authoritative in the archived manifests and split_assignments.csv. Splitting is by example, not by label: the same class may legitimately appear in training, validation, and test sets. Participation, local minibatch order, perturbations, and final adaptation have deterministic derived seed streams. Nonparticipating clients keep their previous private or complete personal state. The campaign enables deterministic PyTorch operations and uses four CPU threads.

### Appendix B.2. Source identities and software

The recorded environment is Python 3.12.13 (conda-forge), PyTorch 2.6.0+cpu, NumPy 2.4.6, and SciPy 1.16.3 on Linux. The code identity recorded in environment.json is

```
ed26f294c161e7b36ca9381097dde07ed5138421d8fc25fe1925211cffa47ccb.
```

This is the runner's normalized code-object hash, not the byte hash of the captured source file. The source and environment records are preserved; metric verification does not execute the captured training code.

The source archives and recorded SHA-256 values are:

• CIFAR-10: https://www.cs.toronto.edu/~kriz/cifar-10-python.tar.gz;

`6d958be074577803d12ecdefd02955f39262c83c16fe9348329d7fe0b5c001ce.`

• Sent140: https://cs.stanford.edu/people/alecmgo/trainingandtestdata.zip;

`004a3772c8a7ff9bbfeb875880f47f0679d93fc63e5cf9cff72d54a8a6162e57.`

The CIFAR loader additionally checks the known source archive MD5. The recorded Sent140 SHA-256 identifies the acquired bytes; no independently supplied expected Sent140 SHA-256 was configured. The dataset-manifest fingerprints are, respectively,

`451d9162dfe51d778bf23b88ddf0f2ac652854e030d3c7251140a27bda924de9`

`446ca84de394f67f82db188945d517a0f96966e298810aef93496ab949b39f06.`

These fingerprints identify the recorded data construction; the audit does not independently establish raw-data authenticity.

## Appendix C. Validation selection and failures

Table C.8. Native validation accuracy (%) at tuning round 20, seed 9001. Each method chooses the best finite candidate. "Nonfinite" denotes an explicitly recorded failure, not an omitted observation. All three ORDERS controls inherit ORDERS' choice.

| Method (R1) | LR 0.01 | LR 0.03 | LR 0.1 |
|---|---|---|---|
| **CIFAR-10** | | | |
| FedAvg-FT | 66.760 | 50.100 | Nonfinite |
| FedPer | 66.140 | 49.000 | Nonfinite |
| FedRep | 58.500 | 49.000 | 49.000 |
| Ditto | 53.980 | 52.320 | Nonfinite |
| ORDERS | 70.880 | 62.760 | Nonfinite |
| **Sent140** | | | |
| FedAvg-FT | 61.292 | 62.121 | 67.431 |
| FedPer | 64.224 | 67.156 | 72.282 |
| FedRep | 61.710 | 64.013 | 69.031 |
| Ditto | 62.097 | 65.934 | 70.198 |
| ORDERS | 63.736 | 67.058 | 72.013 |

The finite-candidate selection leaves 26 completed evaluations and four failed candidates. The failures occur only for the listed CIFAR-10 candidates, not for final runs. The final test results are not substitutions for failed tuning outcomes. Selection on one short tuning seed does not identify a robust optimum over the larger hyperparameter space.

## Appendix D. Complete paired accuracy comparisons

Table D.9 includes all 28 comparisons. The paired differences use unrounded run scores; reported means and intervals are rounded only for display. A positive number favors ORDERS. Probabilities below 0.0001 are reported as $< 0.0001$; other probabilities are rounded to four decimals. The Holm family includes both endpoints and all comparators; no fairness or majority-diagnostic test is included in that family.

Table D.9. Paired ORDERS-minus-comparator accuracy effects (pp), five seeds.

| Comparator (R1) | Mean | Unadjusted 95% CI | Raw p | Holm p |
|---|---|---|---|---|
| **CIFAR-10 / Native** | | | | |
| FedAvg-FT | 3.616 | [2.758, 4.474] | 0.0003 | 0.0073 |
| FedPer | 1.494 | [0.294, 2.694] | 0.0259 | 0.5435 |
| FedRep | 4.188 | [3.637, 4.739] | < 0.0001 | 0.0008 |
| Ditto | 7.404 | [5.676, 9.132] | 0.0003 | 0.0072 |
| ORDERS: uniform | 0.246 | [0.062, 0.430] | 0.0206 | 0.4526 |
| ORDERS: no alignment | 1.606 | [0.115, 3.097] | 0.0403 | 0.7654 |
| ORDERS: no mutation | -0.086 | [-0.235, 0.063] | 0.1853 | 1.0000 |
| **CIFAR-10 / Common FT** | | | | |
| FedAvg-FT | 4.260 | [3.668, 4.852] | < 0.0001 | 0.0010 |
| FedPer | 0.318 | [-0.142, 0.778] | 0.1271 | 1.0000 |
| FedRep | 5.512 | [4.183, 6.841] | 0.0003 | 0.0075 |
| Ditto | 7.342 | [6.383, 8.301] | < 0.0001 | 0.0008 |
| ORDERS: uniform | 0.354 | [-0.268, 0.976] | 0.1895 | 1.0000 |
| ORDERS: no alignment | 0.244 | [-0.330, 0.818] | 0.3035 | 1.0000 |
| ORDERS: no mutation | 0.062 | [-0.032, 0.156] | 0.1407 | 1.0000 |
| **Sent140 / Native** | | | | |
| FedAvg-FT | 0.868 | [0.131, 1.604] | 0.0308 | 0.6151 |
| FedPer | 0.159 | [-0.224, 0.542] | 0.3128 | 1.0000 |
| FedRep | 0.065 | [-0.449, 0.580] | 0.7423 | 1.0000 |
| Ditto | 0.519 | [0.020, 1.017] | 0.0445 | 0.8018 |
| ORDERS: uniform | 0.139 | [-0.122, 0.399] | 0.2136 | 1.0000 |
| ORDERS: no alignment | 0.088 | [-0.186, 0.362] | 0.4245 | 1.0000 |
| ORDERS: no mutation | 0.015 | [-0.105, 0.135] | 0.7505 | 1.0000 |
| **Sent140 / Common FT** | | | | |
| FedAvg-FT | 0.510 | [-0.021, 1.041] | 0.0560 | 0.9514 |
| FedPer | 0.004 | [-0.596, 0.605] | 0.9859 | 1.0000 |
| FedRep | -0.375 | [-1.207, 0.458] | 0.2794 | 1.0000 |
| Ditto | 0.023 | [-0.426, 0.472] | 0.8924 | 1.0000 |
| ORDERS: uniform | 0.166 | [-0.232, 0.564] | 0.3120 | 1.0000 |
| ORDERS: no alignment | 0.139 | [-0.207, 0.486] | 0.3259 | 1.0000 |
| ORDERS: no mutation | 0.022 | [-0.039, 0.084] | 0.3739 | 1.0000 |

## Appendix E. Run-level accuracy

The five columns in Table E.10 are the independent training-seed replicates conditional on the fixed partition. Values for FedAvg-FT are identical across endpoints because the evaluated fine-tuned state is reused, not because two independent evaluations happened to agree. The evidence CSVs preserve full precision, pooled accuracy, client-level outcomes, and state hashes.

Table E.10. Mean client test accuracy (%) by training seed.

| Method (R1) | 101 | 202 | 303 | 404 | 505 |
|---|---|---|---|---|---|
| **CIFAR-10 / Native** | | | | | |
| FedAvg-FT | 76.270 | 76.880 | 76.600 | 77.320 | 77.420 |
| FedPer | 76.730 | 78.590 | 80.290 | 79.690 | 79.800 |
| FedRep | 74.850 | 77.200 | 76.370 | 77.210 | 76.000 |
| Ditto | 73.180 | 73.040 | 73.790 | 74.430 | 71.110 |
| ORDERS | 79.190 | 81.210 | 80.970 | 80.710 | 80.490 |
| ORDERS: uniform | 79.020 | 80.910 | 80.490 | 80.540 | 80.380 |
| ORDERS: no alignment | 76.850 | 78.620 | 81.150 | 78.340 | 79.580 |
| ORDERS: no mutation | 79.240 | 81.380 | 80.860 | 80.860 | 80.660 |
| **CIFAR-10 / Common FT** | | | | | |
| FedAvg-FT | 76.270 | 76.880 | 76.600 | 77.320 | 77.420 |
| FedPer | 79.760 | 80.870 | 80.950 | 81.400 | 81.220 |
| FedRep | 73.090 | 75.010 | 76.910 | 77.220 | 76.000 |
| Ditto | 72.830 | 72.920 | 75.280 | 74.850 | 73.200 |
| ORDERS | 80.340 | 80.560 | 81.500 | 81.890 | 81.500 |
| ORDERS: uniform | 80.260 | 80.420 | 81.710 | 80.930 | 80.700 |
| ORDERS: no alignment | 79.580 | 80.870 | 80.880 | 81.610 | 81.630 |
| ORDERS: no mutation | 80.200 | 80.600 | 81.480 | 81.760 | 81.440 |
| **Sent140 / Native** | | | | | |
| FedAvg-FT | 73.372 | 73.609 | 73.682 | 74.441 | 74.108 |
| FedPer | 74.069 | 74.769 | 74.733 | 74.728 | 74.455 |
| FedRep | 74.670 | 74.847 | 74.319 | 75.073 | 74.315 |
| Ditto | 74.280 | 74.103 | 74.376 | 74.314 | 73.884 |
| ORDERS | 74.300 | 75.067 | 74.998 | 75.112 | 74.072 |
| ORDERS: uniform | 74.126 | 74.819 | 75.051 | 74.701 | 74.161 |
| ORDERS: no alignment | 74.004 | 74.999 | 74.722 | 75.064 | 74.322 |
| ORDERS: no mutation | 74.300 | 74.956 | 75.110 | 75.150 | 73.961 |
| **Sent140 / Common FT** | | | | | |
| FedAvg-FT | 73.372 | 73.609 | 73.682 | 74.441 | 74.108 |
| FedPer | 73.971 | 74.574 | 74.135 | 74.527 | 74.535 |
| FedRep | 74.745 | 75.378 | 74.538 | 74.487 | 74.488 |
| Ditto | 74.305 | 74.664 | 74.085 | 74.631 | 73.961 |
| ORDERS | 74.083 | 74.463 | 73.963 | 75.280 | 73.975 |
| ORDERS: uniform | 73.825 | 74.781 | 73.929 | 74.769 | 73.631 |
| ORDERS: no alignment | 73.528 | 74.449 | 73.698 | 75.246 | 74.145 |
| ORDERS: no mutation | 74.083 | 74.463 | 73.852 | 75.280 | 73.975 |

## Appendix F. Evidence audit and claim changes

The supplied archive has SHA-256

```
b7e049a8fd3ea90ded4a0bbcd504364c46c076c1eeaf60b7699236ef7a431a69.
```

All ZIP CRC checks and 9,438 included-file checksums pass. The archive contains 9,439 members, 80 final run records, 8,000 final training-round records, and 160 final evaluation endpoints. Independent parsing recomputes metrics from 973,275 saved prediction records, including tuning and duplicated endpoint records; this count does not represent unique test examples. Recomputed summaries agree with the saved aggregate results. The audit checks recorded participant schedules, coefficient assignments, parameter-payload accounting, split consistency, and completed/failed tuning outcomes. It never executes the archived training source.

The record check is limited by the supplied archive. Its exclusion inventory lists 860 model files totaling 3,325,439,029 bytes. Model inference cannot be independently replayed from those absent files, and the audit did not reacquire raw images or tweets. Hash consistency is evidence of internal record identity, not independent proof of an experimental outcome.

The revision replaces earlier, unaudited accuracy summaries rather than combining them with the R1 campaign. It also removes unsupported global-to-local gain columns, the 0.5× communication interpretation, speech and attack results not present in this evidence, the curvature– conflict guarantee, and an unconditional convergence-rate claim. No "55-point" or historical 19.7-point personalization gain is used as a result of this campaign. Cross-method differences and component-removal effects are stated in percentage points, with their comparator and endpoint identified.

The empirical study reports complete measured ablation tables. It does not imply completion of unrecorded baseline, attack, architecture, or hyperparameter-sweep experiments.